\documentclass{article}
\usepackage{arxiv}
\usepackage[utf8]{inputenc} 
\usepackage{hyperref}       
\usepackage{booktabs}       
\usepackage{amsfonts}       
\usepackage{nicefrac}       
\usepackage{microtype}      
\usepackage{lipsum}
\usepackage{graphicx}
\usepackage{amssymb}
\usepackage{amsmath}

\usepackage{url}
\usepackage{algorithm2e}
\usepackage{xspace}
\usepackage{tikz}
\usepackage{multirow}
\usepackage{hyperref}

\newcommand{\drx}{LXDR\xspace}
\usepackage[font=small,labelfont=bf]{caption}

\title{Local Explanation of Dimensionality Reduction}

\author{
  Avraam Bardos\\
  Aristotle University of \\Thessaloniki, 54636, Greece\\
    \texttt{ampardos@csd.auth.gr}\\
     \And
     Ioannis Mollas\\
  Aristotle University of \\Thessaloniki, 54636, Greece\\
    \texttt{iamollas@csd.auth.gr}\\
     \And
       Nick Bassiliades\\
  Aristotle University of \\Thessaloniki, 54636, Greece\\
             \texttt{nbassili@csd.auth.gr}\\
             \And
               Grigorios Tsoumakas\\
  Aristotle University of \\Thessaloniki, 54636, Greece\\
             \texttt{greg@csd.auth.gr}\\
}

\begin{document}
\maketitle

\begin{abstract}
Dimensionality reduction (DR) is a popular method for preparing and analyzing high-dimensional data. Reduced data representations are less computationally intensive and easier to manage and visualize, while retaining a significant percentage of their original information. Aside from these advantages, these reduced representations can be difficult or impossible to interpret in most circumstances, especially when the DR approach does not provide further information about which features of the original space led to their construction. This problem is addressed by Interpretable Machine Learning, a subfield of Explainable Artificial Intelligence that addresses the opacity of machine learning models. However, current research on Interpretable Machine Learning has been focused on supervised tasks, leaving unsupervised tasks like Dimensionality Reduction unexplored. In this paper, we introduce \drx, a technique capable of providing local interpretations of the output of DR techniques. Experiment results and two \drx use case examples are presented to evaluate its usefulness.
\end{abstract}


\section{Introduction}

In recent years, the amount of data being produced on a regular basis, either through sensors or other technological means, has been rapidly increasing. Building on top of the availability of abundant data, Artificial Intelligence (AI) and Machine Learning (ML) have become an integral part of our daily lives. End-to-end transparency of AI and ML algorithms is crucial in high-risk applications such as disease diagnosis or criminal justice tools. Explainable AI (XAI) and Interpretable ML (IML) are fields that are researching how to make decision-making systems interpretable~\cite{XAI}.

The data that are used for training AI systems are often high-dimensional, comprising thousands of features. This raises the computational complexity of building such systems on the one hand, while it also hurts their effectiveness on the other, as some  features can be irrelevant, noisy or redundant. Dimensionality Reduction (DR) is a key solution for addressing these issues~\cite{dr_survey}. However, non-linear DR methods are opaque. The features extracted by such DR methods are difficult, if not impossible, to comprehend. By using them to solve supervised tasks, such as classification or regression, we are not able to identify which original feature contributed the most to a prediction.

Limited research has been conducted to address the interpretability of DR techniques. Related work includes the use of subspace projections~\cite{d2}, as well as adjusting existing methods designed to explain predictive models, such as LIME~\cite{tsne1}. 
None of these methods however is general enough to be applicable to any DR technique. To fill this gap, we present Local eXplanation of Dimensionality Reduction (\drx), a model-agnostic technique for explaining the results of any non-linear DR technique, with the assumption that this technique will be able to reduce new instances. \drx extracts interpretations via a local surrogate DR model, obtained by fitting a linear multi-target prediction model in the neighborhood of a reduced instance. Empirical results verify the interpretability capabilities of \drx.

The rest of this paper is organized as follows. Section~\ref{sec:back} presents background material and related work on interpretable dimensionality reduction. Section~\ref{sc:drx} presents \drx and Section~\ref{sec:exp} our experiments. Finally, Section~\ref{sec:con} presents the conclusions of this work, as well as future research directions.  
\section{Background and Related Work}
\label{sec:back}

This section starts by reviewing popular DR techniques. It then gives a brief introduction to the topic of interpretability, with a focus on surrogate models. Finally, it discusses related work on interpretable dimensionality reduction.

\subsection{Dimensionality Reduction}
A key use of DR is for enabling the visualization of high-dimensional datasets, as humans cannot comprehend a lot of dimensions. In addition, DR is beneficial when training ML models for supervised tasks, through the removal of noisy or redundant features that delay training and may even hurt generalization. 

One of the most popular DR algorithms is Principal Component Analysis (PCA)~\cite{PCA}. Aiming to retain as much information from the data as possible through optimization, PCA constructs a reduced number of features, dubbed components, expressing linear relations among the original features. 
A variation of PCA is Sparse Principal Components Analysis (SPCA), which involves restricting as many coefficients as possible to being zero (sparsity constraint)~\cite{spca}. In the same vein, SCoTLASS combines PCA and least absolute shrinkage and selection operator (LASSO) to produce sparser models~\cite{d3}.

Non-Negative Matrix Factorization (NMF) is a technique capable of both dimensionality reduction, and topic extraction, applicable to data with non-negative values~\cite{NFM}. Linear Discriminant Analysis (LDA)~\cite{xanthopoulos2013linear} attempts to project the data into components that maximize class separation. LDA is a supervised technique that requires ground truth information on the target (class or labels), unlike PCA and NMF.

Kernel PCA (KPCA)~\cite{KPCA} is a variation of PCA that allows the representation of non-linear relations among the original features through the use of a kernel function. The t-distributed Stochastic Neighbor Embedding (t-SNE)~\cite{TSNE} algorithm was designed for data visualization. Each high-dimensional data sample is represented by a two- or three-dimensional point in t-SNE, with a high likelihood of similar (dissimilar) samples being modelled by adjacent (distant) points. Other non-linear reduction techniques are Isometric mapping (IsoMap)~\cite{isomap} and Spectral Embedding~\cite{SP}. Finally, neural Autoencoders (AE) are also used for dimensionality reduction~\cite{AutoEn}.


\subsection{Interpretability}
\label{IMLearning}
Interpretability concerns the ability of ML models to provide reasons for their outputs. However, not every model is able to provide such information right out of the box. Such models are called \textit{black box} models. As a result, a large body of research has focused on developing techniques that enable the interpretability of black box models. 

Interpretability techniques can be either: a) global, offering explanations for the entire structure of the model, b) local, providing interpretations for a model's output regarding a specific instance, or c) both. Furthermore, interpretability techniques can be classified as model-agnostic, applicable to any type of ML model, or model-specific, designed to interpret particular models and architectures~\cite{XAI}.

A simple technique to interpret a black box is to use another, transparent model to explain it. The latter is called a {\em surrogate model}, and learns the data and the complex model's predictions with the ultimate goal of extracting explanations. LIME, a state-of-the-art interpretability technique, follows this paradigm, but in local subspaces~\cite{lime}. To interpret an instance's prediction, LIME creates a synthetic local neighborhood around the instance. After using the black box to predict the generated neighbors, it trains a linear model on the neighborhood and the predictions, to extract weights for the features. These weights express the influence of a feature on the prediction of the instance, which is the final interpretation.

\subsection{Interpretable Dimensionality Reduction}

\label{sec:idr}

\begin{figure}[ht]
\centerline{\includegraphics[width=0.8\textwidth]{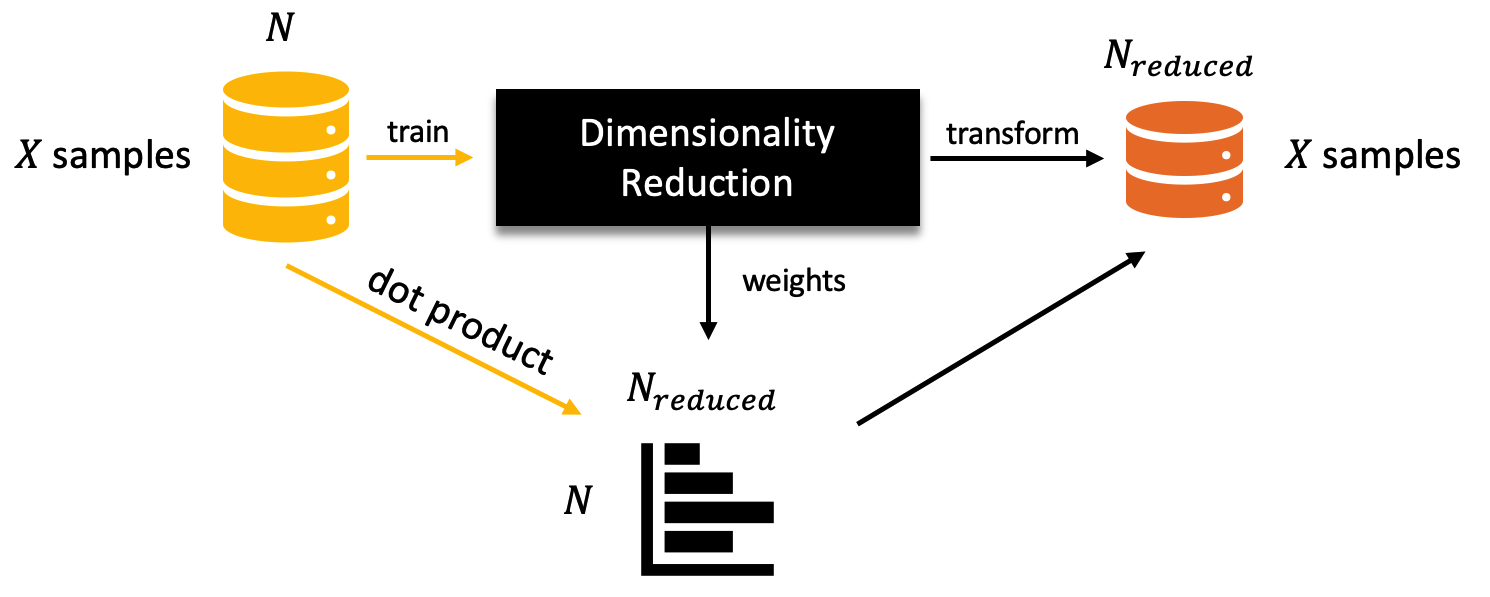}}
\caption{Reduced space data extraction through a combination of weights and original data}
\label{fig:TransparentPCA}
\end{figure}

Because of their linear nature, PCA, NMF and LDA are intrinsically interpretable. As shown in Figure \ref{fig:TransparentPCA}, internally, PCA learns a matrix of weights that linearly map the original dimensions $N$ to the reduced dimensions $N_{reduced}$. The reduced data can therefore be obtained through a multiplication of the original data with this matrix. 

As an example, Table~\ref{tab:weights} shows the matrix of weights obtained when reducing the 4 dimensions of the Iris dataset~\cite{Dua2019} to 3 dimensions via PCA. These weights constitute an interpretation of the PCA transformation. We can see, for example, that among the original features, $F_3$ influences more the first component, and $F_2$ the second and third components. 


\begin{table}[ht]
\caption{Components' weights extracted via PCA in the Iris dataset}
\centering
\begin{tabular}{rcccc}
\hline
\textbf{Dimensions} & \textbf{$F_1$} & \textbf{$F_2$} & \textbf{$F_3$} & \textbf{$F_4$} \\ \hline
$1^{st}$ Component       & 0.361       & -0.084      & 0.856       & 0.358       \\ 
$2^{nd}$ Component       & 0.656       & 0.73        & -0.173      & -0.075      \\ 
$3^{rd}$ Component       & -0.582      & 0.597       & 0.076       & 0.545       \\ \hline
\end{tabular}
\label{tab:weights}
\end{table}

These weights, however, are not provided by non-linear DR approaches. KPCA is uninterpretable when radial basis function or polynomial kernels are used. One solution, addressing the opaqueness of KPCA~\cite{d2}, employs a subspace on which features are projected before they are mapped into a high dimensional space using a kernel function. Hence, the mapping to the original features remains linear via a projection matrix, making kernel DR techniques interpretable. This technique, however, interferes with the algorithm of KPCA, which has a negative impact on the technique's performance. In addition, there is no publicly available implementation to experiment with it.

The non-parametric non-linear mapping of t-SNE is also difficult to interpret. Given that t-SNE uses neighborhoods to generate new representations for instances, known as embeddings, LIME was considered as a good fit for improving its interpretability in~\cite{tsne1}. However, as LIME was designed for supervised models, a number of adjustments were explored. To begin, a LIME-inspired neighborhood creation process based on SMOTE was used. Afterwards, the neighbors were projected onto the reduced space. Lastly, a LASSO regression model was utilized to explain an instance's projection. Unfortunately, this work does not give details on how the second step is performed. This is an important issue, considering that t-SNE cannot project new samples and must create all the embeddings from scratch each time. 

Manifold learning techniques, including IsoMap and Spectral Embedding, are also uninterpretable. Finally, an AE model can be interpreted intrinsically if it only contains an input layer, a latent layer, and an output layer. However, if it incorporates additional hidden layers, usually to increase the capacity, it becomes uninterpretable. 





\begin{figure*}[htb]
\centerline{\includegraphics[width=1\textwidth]{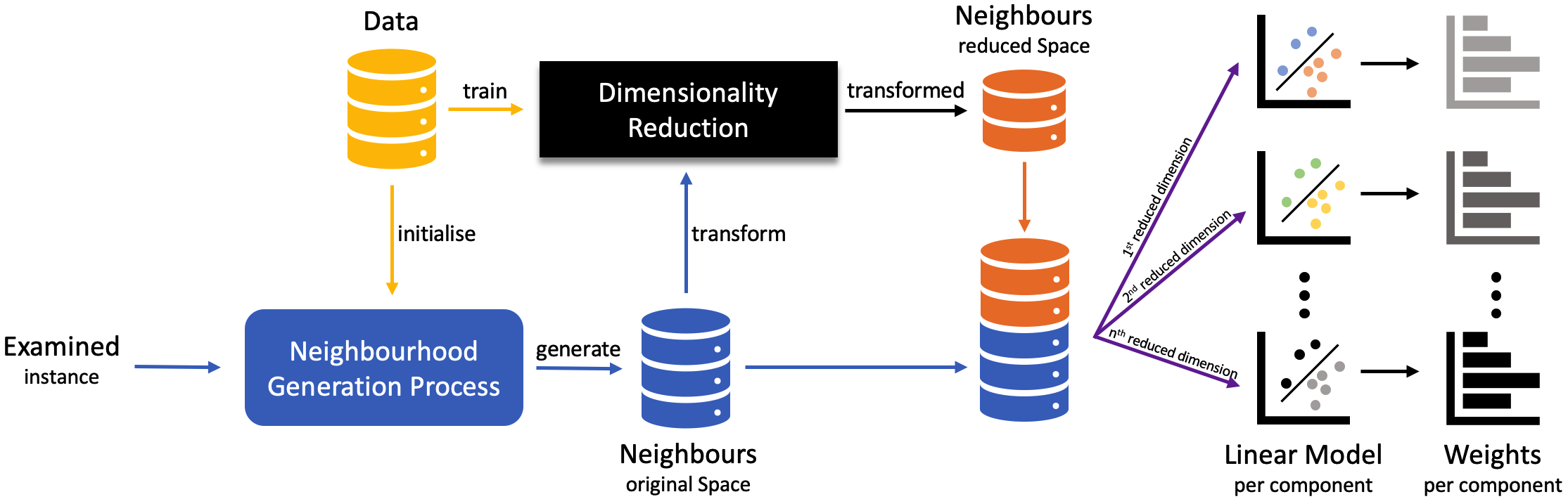}}
\caption{Workflow approach for the dimensionality reduction interpretation}
\label{fig:drx}
\end{figure*}

\section{Our Approach}
\label{sc:drx}

Given a black box DR model
, the main idea of \drx is to build a linear, hence interpretable, surrogate of that model. Instead of building a single global linear surrogate, \drx follows a local approach, building a surrogate DR model around each particular instance for which an explanation of the black box model is needed. We consider two such uses cases: a) explaining decisions of supervised models operating on data that have been reduced by black box DR models (see Figure~\ref{fig:usecase1}), and b) explaining outlier instances in the reduced space (see Figure~\ref{fig:outlier}) of a black box DR model. 

To achieve its purpose, \drx trains one local linear regression model per reduced dimension, using the original dimensions as predictor variables. The output of \drx is therefore a matrix with the parameters of these linear models, representing the contribution of each original dimensions towards each of the reduced dimensions. This matrix can be used to compute a set of feature weights in the original space, through the inverse transform function as discussed later in Section~\ref{sec:ucr}, given a set of feature weights in the reduced space for the first use case. Moreover, utilizing \drx to obtain insight on which original dimensions contribute more to the outlierness of an instance in the reduced space is examined in the second use case. The local nature of \drx is expected to lead to more faithful interpretations of non-linear black box DR models, compared to a single global linear surrogate. \drx is model-agnostic, as it can be used in tandem with any black box DR model. Figure \ref{fig:drx} presents the workflow of \drx, while its pseudocode is given in Algorithm~\ref{alg:drx}. The rest of this section describes \drx in more detail.

\begin{algorithm}[htb]
\SetKwInOut{Input}{Input}\SetKwInOut{Output}{Output}
\Input{Instance $x_i$,
Original Dataset $D$,\\
Trained DR\ $r$, \# of Neighbors $K$,\\
Neighbourhood Generation Technique $NG$}
\Output{Coefficients $weights$}
    $weights \gets \emptyset$ \\
    \# Generate a neighborhood for the given instance\\
    $B \gets$ $x_i\cup NG$($D$, $K$, $x_i$)\\
    \# Apply DR to the neighbors \\
    $B' \gets r(B)$\\
    \# For each dimension (component)\\
    \For{$dimension$ in $reduced\_neighbors$}{
        \# Take the values of the selected dimension of the neighbors\\
        $rd \gets B'[dimension]$\\
        \# Train a linear model to predict those values\\
        $model \gets train\_linear\_model(B, rd)$\\
        \# Extract the weights as explanation for this component\\
        $weights \gets weights \cup [model.get\_weights]$
    }
 \textbf{return} $weights$
 \caption{The pseudocode of \drx}
 \label{alg:drx}
\end{algorithm}

We assume that the given black box model can be used to reduce an arbitrary instance from $N$ dimensions to $N_r$, as we will use it to reduce the given instance $x$ as well as a set of nearby instances. Note that some DR techniques, like t-SNE for example, do not offer this functionality. They act similarly to transductive models in supervised learning, reducing only the data they have been trained on. Retraining them on the original data expanded with the given instance and its neighbors would lead to a different DR model than the one we aim to approximate via a surrogate. 



\begin{figure}[ht]
\centerline{\includegraphics[width=0.55\textwidth]{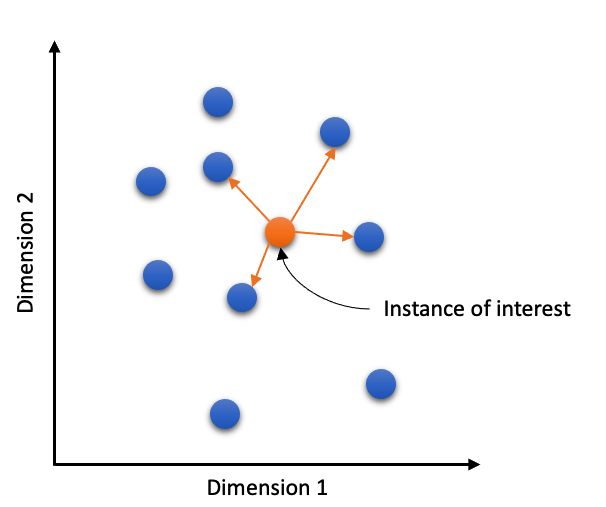}}
\caption{Instance neighborhood generation}
\label{fig:9}
\end{figure}

A crucial component of our technique is the Neighborhood Generation process $NG$. For a given instance $x_i$, we use an $NG$ technique (Figure~\ref{fig:9}), in order to find or generate, depending on the technique, similar instances (neighbors). This utilization of a neighborhood, gives to our technique its local flavor. 

For the neighborhood generation, we investigated two techniques. The first one is a very simple and commonly used technique (KNNs). The second one is LIME's $NG$ technique for tabular data, as presented in Section~\ref{sec:back}.

KNNs algorithm relies on the assumption that the most similar instances are close as well. Manhattan and Euclidean are the most often used metrics for calculating the distance between instances. The Euclidean distance was applied in our case. As inputs, this technique requires both the learned DR technique and the whole training dataset.

LIME was considered as a second $NG$ technique. LIME was presented in Section~\ref{IMLearning} as a local model-agnostic interpretation technique that uses surrogate models to explain black box models. LIME produces these explanations by generating new synthetic data instances around a given instance. LIME does not directly provide us with these neighborhoods, and therefore, we made adjustments to its implementation to access the generated neighbors. LIME is a better option when only a portion of the training data is available, and it is also faster than KNNs.



Therefore, applying an $NG$ technique, we are gathering the neighbors $n^j$ of an instance $x_i$, in a set $B = \{x_i, n^1, n^2, \dots, n^K\}$ containing the given instance $x$ and its neighbors $n^{(i)}$, $i=1 \ldots K$. 
However, in order to train our linear model, we must also have the target values, which in this case are the transformed representations of the neighbors. Therefore, we reduce the data set $B$ using the black box DR model $r$: $B' \leftarrow r(B)$. 

Data sets $B$ and $B'$ form a multi-target regression task, where $B$ are the input values and $B'$ the output values. To build the surrogate DR model, we simply train one linear regression model for each output variable, i.e. for each reduced dimension. We leave more sophisticated multi-target regression approaches \cite{Tsoumakas2014,Tsoumakas2016} for future work. However, we do employ a weighting mechanism, so that the linear models pay more attention to neighbors closer to the given instance $x$, as well as to $x$ itself. Let $d(x,n^{(i)})$ denote the Euclidean distance of instance $x$ from a neighbor $n^{(i)}$. The given instance takes a weight of 1, while each neighbor $n^{i}$ is weighted with a weight $w^{(i)}$ given via the following equation:

\begin{equation}
    \label{eq:weighting}
    w^{(i)} = e^{-2\cdot d(x,n^{(i)})} 
\end{equation}

Finally, \drx returns a matrix $weights \in \mathbf{R}^{N\times N_r}$ comprising the parameters learned by the linear models. 


\section{Experiments}
\label{sec:exp}
This section focuses on the basic parameters of our technique, the setup, the quantitative experiments, and the qualitative experiments. The experiments' code will be publicly available on the \drx's GitHub repo: ( \url{https://github.com/avrambardas/Interpretable-Unsupervised-Learning.git}).

In the setup section, we introduce the datasets utilized in the experiments, as well as the features we take into account for each one. The parameters of our technique and how they are employed, as well as the metrics used to evaluate \drx, are also presented. In the quantitative experiments section, we provide all the results supporting the effectiveness of our technique for the various datasets. We also perform a scalability study regarding the time performance of \drx. Finally, the qualitative experiments' section demonstrates the usefulness of our technique on the component explanation in two different use cases.

\subsection{Setup}

We experimented with three datasets, which demonstrate the universality of our technique to both regression and classification tasks. Therefore, two classification related datasets (Iris\footnote{\url{https://bit.ly/3rMKxpA}}, Digits\footnote{\url{https://bit.ly/3jr6h6P}}) and one regression related (Diabetes\footnote{\url{https://bit.ly/3xfuTnS}}) were considered. More information can be found in Table~\ref{tab:datasets}.

\begin{table}[ht]
\caption{Information about datasets in terms of size, features, task, and classes}
\centering
\begin{tabular}{lrrrl}
\hline
\textbf{Dataset} & \textbf{Instances} & \textbf{Features} & \textbf{Task} & \textbf{Classes}             \\ \hline
Iris             & 150                      & 4                       & Classification                      & 3 \\ 
Digits           & 1797                      & 64                      & Classification                     & 10 \\ 
Diabetes         & 442                      & 10                      & Regression                     & Regression                \\ \hline
\end{tabular}
\label{tab:datasets}
\end{table}

We applied DR to each dataset, trying to preserve at least 95\% of the original information. Iris, for example, was reduced from 4 to 3 dimensions, Diabetes from 10 to 8 dimensions, and Digits from 64 to 25 dimensions. Furthermore, only 25\% (449 out of 1,797 instances) of the entire data was used for Digits, in order to achieve faster execution time.

Additional parameters were introduced and employed for the experiments to improve the results, boosting our technique to more accurate results with fewer mistakes. These parameters, as well as their descriptions, are presented below:

\begin{itemize}
    \item [\textbf{K:}] K is the \# of neighbors, and is set by the user, default = $10\%$ of the total instances.
    \item [\textbf{NG:}] NG can be either LIME or KNNs, default = KNNs.
    \item [\textbf{AA:}] Auto Alpha (AA) is the linear model's regularization strength (alpha). If AA is true, our approach would select the optimum alpha for the specific instance by experimenting with several alpha values. The selection is based on a self-evaluation procedure, which measures the performance of the linear model to the original data. Otherwise, if AA is false, the linear model's default value of alpha is utilized.
\end{itemize}

\subsubsection{Evaluation Metrics}
We employ two different evaluation methods to assess the performance of our technique. The first one, \textit{instance difference}, uses the extracted by \drx weights $weights_{\drx}$ to reduce the dimensions of an instance $x$ and compare the reduced representation $x'_{\drx}=x\cdot weights_{\drx}$, with the reduced representation of the instance as provided by the DR technique $x'=r(x)$. The second metric, \textit{weights difference}, assesses how much different the weights $weights_{\drx}$ produced by \drx are from the original weights of a DR technique $weights$. However, this metric is applicable only to intrinsically interpretable DR techniques.


\begin{align*}
ED(y, \hat{y})&=  \sqrt{\sum_{i=1}^{n}(y_i - \hat{y_i})^2}
\end{align*}

For these two metrics, we measure the Euclidean Distance (ED). In more depth, ED calculates the distance between two points. We consider that $y$ and $\hat{y}$ correspond to $x'$ and $x'_{\drx}$ in the first metric, while $weights$ and $weights_{\drx}$ in the second.

\subsection{Quantitative Experiments} \label{quantitativeExp}

In this section, we evaluate the performance of \drx. We compare \drx to a global variation of \drx, GXDR, which does not employ neighbors but instead is trained on the whole dataset. Three DR techniques, PCA, RBF KPCA, and AE, are utilized in the experiments to demonstrate our approach's applicability and generalizability to a variety of dimensionality reduction techniques. In the first set of experiments, we measure the \textit{weights difference} between the weights provided by \drx and GXDR applied in PCA, which is interpretable, and we can extract the original weights. The second set of experiments compare the performance of \drx and GXDR based on the \textit{instance difference} using RBF KPCA and AE as targeted DR techniques. We measure ED for both metrics across the three datasets. 

The \drx parameters we use are $AA=True$, $NG=KNNs$, and $K=\{50, 150, 750\}$, for the Iris, Diabetes, and Digits datasets, respectively. In our evaluations, the KNNs $NG$ technique outperformed the LIME technique, in terms of the aforementioned metrics. As a result, KNNs $NG$ is used in the experiments presented below. However, LIME is still a good alternative, particularly when the training dataset is unavailable or just partially available. Finally, we generate 25 synthetically generated datasets with 1K samples and $\{|F|=10l | l \in \mathbf{N}, n \leq 25\}$ number of features, to measure the weights difference and instance difference of \drx with different number of neighbors $K = \{|F|/1000,2|F|/1000,3|F|/1000\}$, as well as to measure its time performance.

\subsubsection{Weights Difference}

We measure the average \textit{weights difference} between the weights produced by \drx for each instance and the interpretable DR approach (PCA) in this set of experiments. We also compare \drx's performance with that of its global variant, GXDR. In Table~\ref{tab:wd-exp}, we compare \drx and GXDR, and we see that \drx outperforms its variation in all cases based on ED. We can also see that the performance difference is significant in two of the three examples.

\begin{table}[ht]
\caption{\drx and GXDR performance based on the \textit{weights difference} metric across datasets. For visualization purposes, all values have been multiplied by 100}
\centering
\begin{tabular}{rcc}
\hline
\textbf{}        & \multicolumn{2}{c}{\textbf{ED}}  \\ \hline
\textbf{Dataset} & \textbf{\drx}  & \textbf{GXDR}  \\ \hline
Iris             & \textbf{0.9743} & 80.7851       \\
Diabetes         & \textbf{0.1669} & 9.7474        \\
Digits           & \textbf{5.3172} & 5.3961        \\\hline
\end{tabular}
\label{tab:wd-exp}
\end{table}

\begin{figure}[ht]
\centerline{\includegraphics[width=0.5\textwidth]{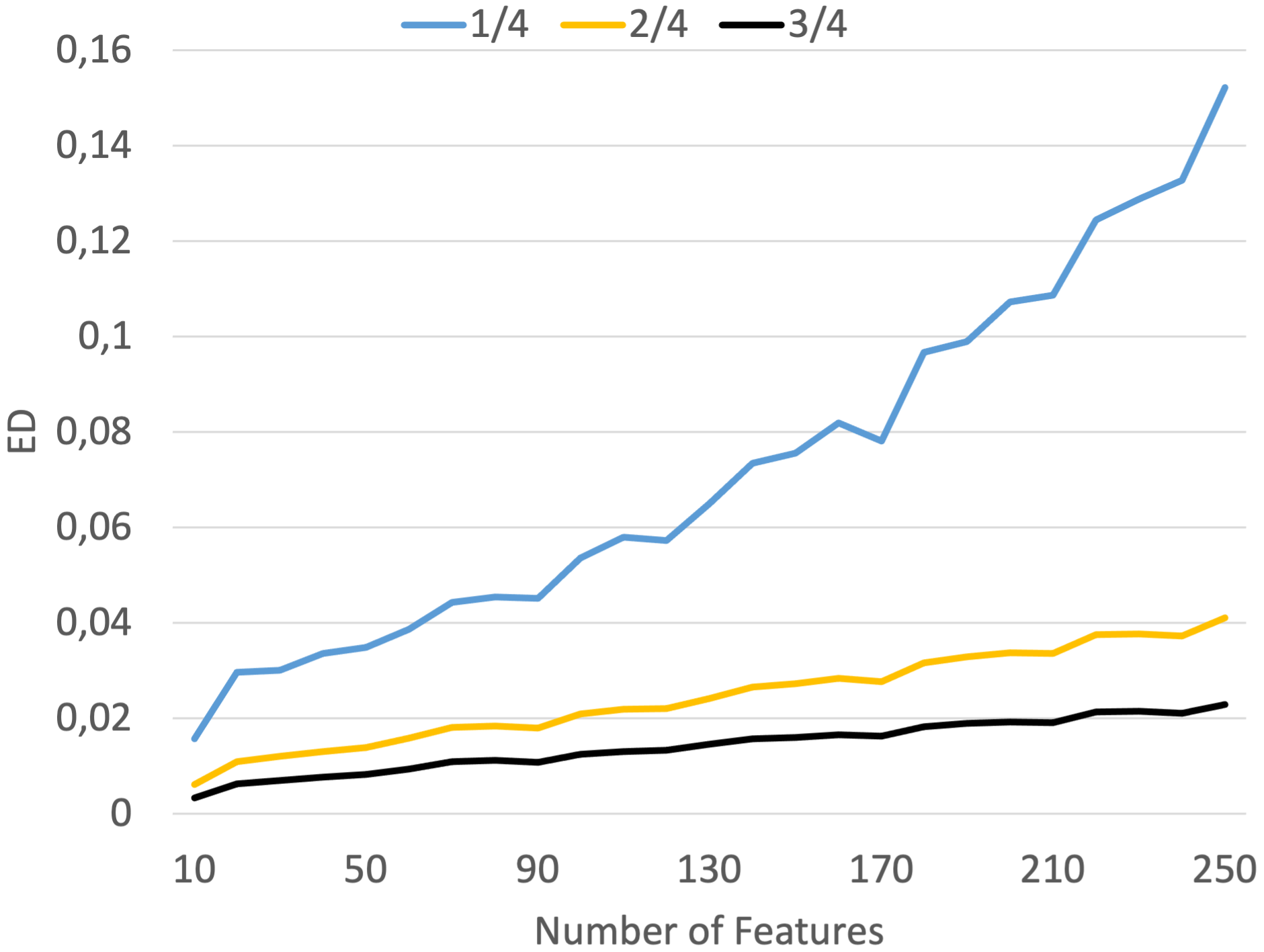}}
\caption{\drx performance based on \textit{weights difference} on synthetic datasets for three distinct \textit{number of neighbors} choices. For visualization purposes, all values have been multiplied by 100}
\label{fig:wdg}
\end{figure}

We also evaluate \drx performance using the synthetic datasets we produced based on this metric. Figure~\ref{fig:wdg} depicts \drx with different values for $K$ to perform in these datasets. Using greater numbers in $K$, such as $1000\cdot3/4=750$, clearly resulted in more accurate weights. However, it is still local because \drx weights the neighbors as shown in Eq.~\ref{eq:weighting}, and this is why it outperforms its global counterpart. GXDR performance in this experiment ranged from $ED = 0.029\cdot100=2.9$ to $ED = 0.204\cdot100=20.4$, for 10 to 250 features, respectively, and was not included in the figure for visualization purposes.

\begin{table}[ht]
\caption{\drx and GXDR performance based on the \textit{instance difference} metric across datasets. For visualization purposes, all values have been multiplied by 100}
\centering
\begin{tabular}{rcll}
\hline
\textbf{}                 & \textbf{}             & \multicolumn{2}{c}{\textbf{ED}}   \\ \hline
\textbf{Dataset}          & \textbf{DR} & \multicolumn{1}{c}{\textbf{\drx}} & \multicolumn{1}{c}{\textbf{GXDR}} \\ \hline
\multirow{2}{*}{Iris}     & KPCA                  & \textbf{3.9099}                    & 7.1388                        \\
                          & AE                    & \textbf{4.9070}                    & 8.3904    \\
\multirow{2}{*}{Diabetes} & KPCA                  & \textbf{2.0346}                    & 8.0876                        \\
                          & AE                    & \textbf{7.8590}                    & 10.6608   \\
\multirow{2}{*}{Digits}   & KPCA                  & \textbf{3.9550}                    & 4.4079                        \\
                          & AE                    & \textbf{29.4543}                   & 39.4562   \\ \hline
\end{tabular}%
\label{tab:id-exp}
\end{table}

\subsubsection{Instance Difference}

In this set of experiments, we measure the performance of \drx and GXDR, in terms of instance difference, on reducing instances based on the weights produced by these techniques. In Table~\ref{tab:id-exp}, we notice that \drx outperforms GXDR in every dataset. An interesting observation here is that both techniques produce better results when interpreting the output of KPCA in contrast to AE. This can be due to the higher complexity of the AE model.

\begin{figure}[ht]
\centerline{\includegraphics[width=0.5\textwidth]{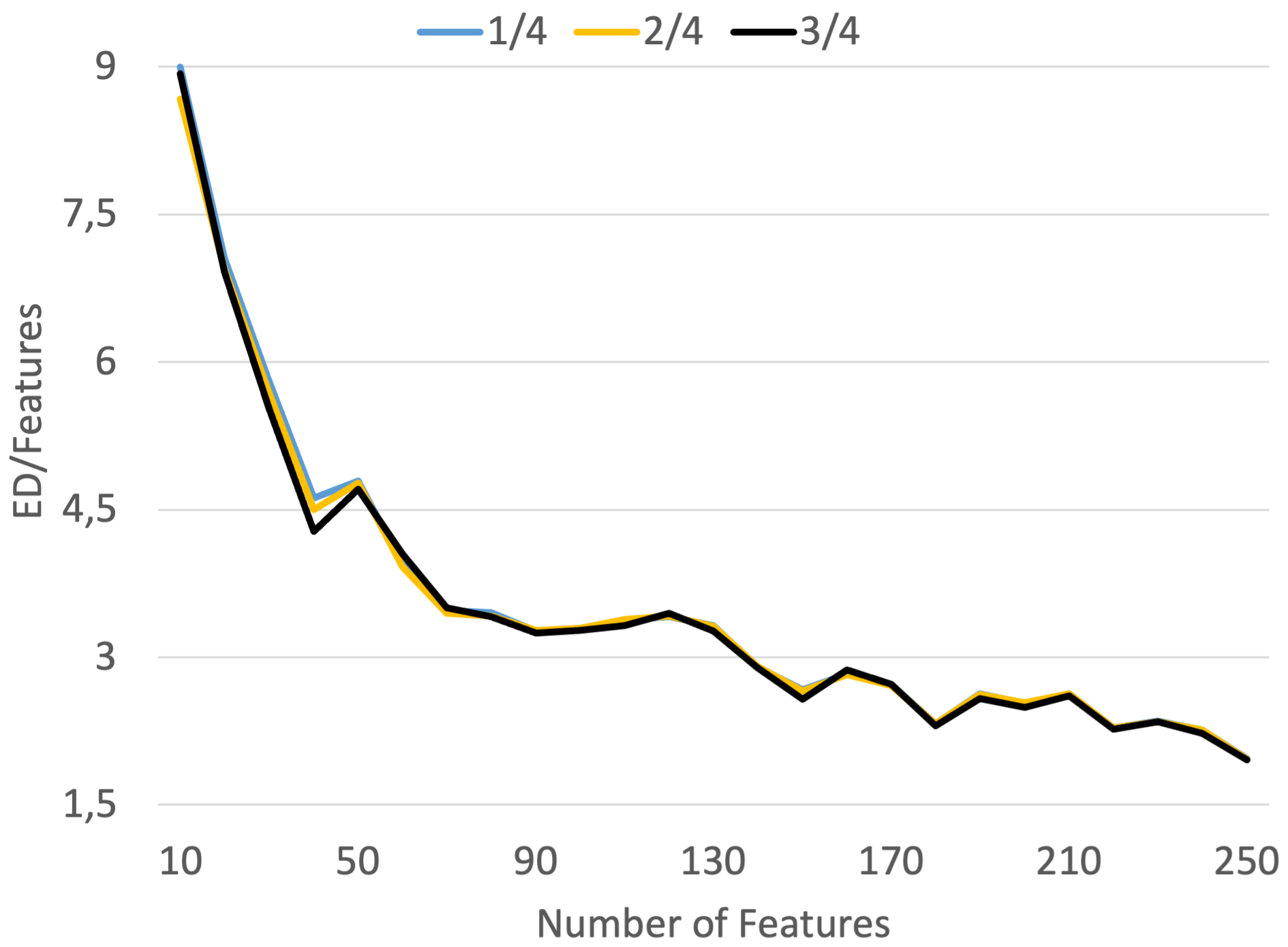}}
\caption{\drx performance based on \textit{instance difference} on synthetic datasets for three distinct \textit{number of neighbors} choices. For visualization purposes, all values have been multiplied by 100}
\label{fig:idg}
\end{figure}

Furthermore, we assess \drx's performance on this metric using the synthetic datasets. Figure~\ref{fig:idg} illustrates how \drx performs on these datasets with varying $K$ values. In this metric, it is quite ambiguous which number of neighbors is preferable, but $1000\cdot3/4=750$ appears to be the best option. In particular, the average performance across the datasets is $1000\cdot1/4=3.56$, $1000\cdot2/4=3.52$, and	$1000\cdot3/4=3.51$, while GXDR has a mean value of $3.55$.

\begin{figure}[ht]
\centerline{\includegraphics[width=0.5\textwidth]{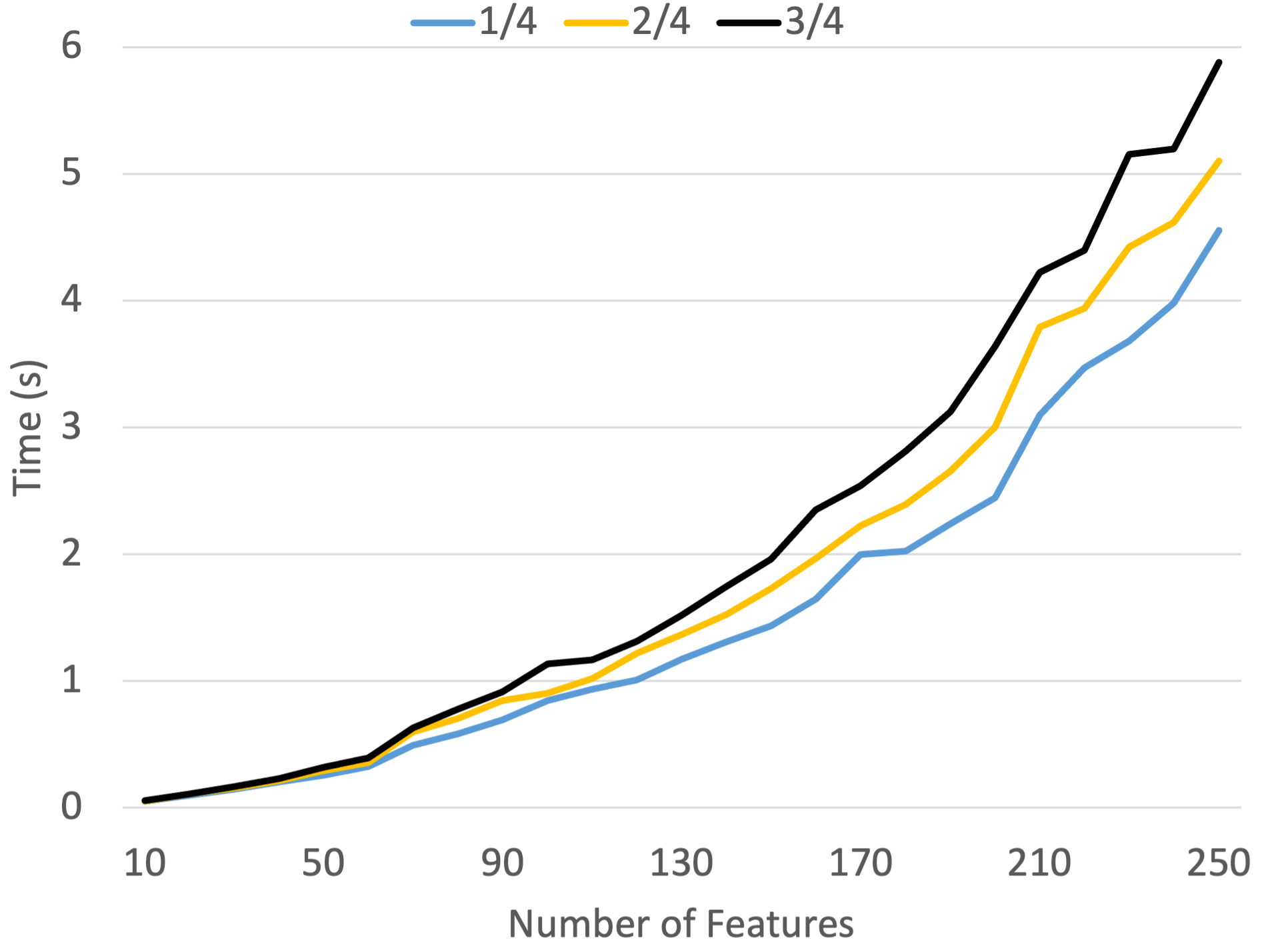}}
\caption{\drx \textit{time performance} on synthetic datasets for three distinct \textit{number of neighbors} choices}
\label{fig:scalability}
\end{figure}

\subsubsection{Scalability}
The final experiment in this section examines the \drx's time performance. Figure~\ref{fig:scalability} shows the average time required by \drx to generate an explanation for an instance for different values of K across the datasets. There is no noticeable difference in \drx's time performance for datasets with fewer features. However, with higher-dimensional datasets, we can see the technique's response time increases as the number of neighbors increases. In the dataset with 250 features, for example, there is a 1-second difference between the three K alternatives. However, we did not optimize or parallelize this process, thus further optimization would result in improved time performance.

\begin{figure*}[ht]
\centerline{\includegraphics[width=0.75\textwidth]{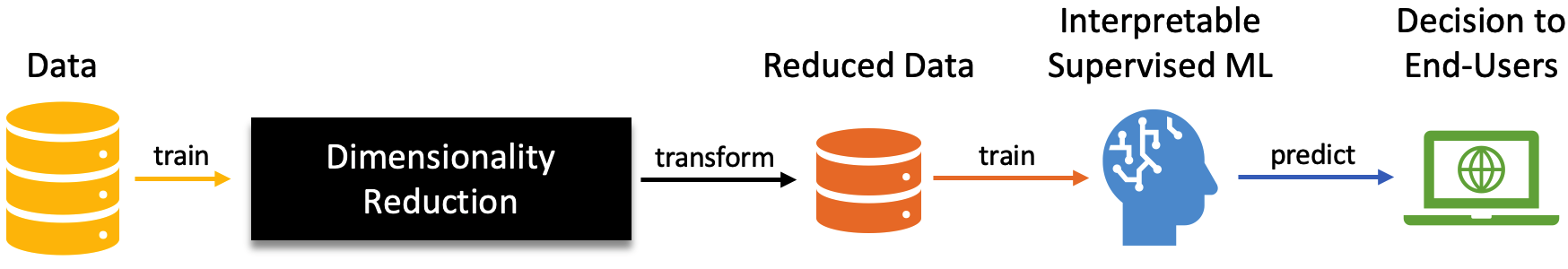}}
\caption{This workflow illustrates a pipeline containing a supervised decision-making system that makes use of data processed by a DR approach}
\label{fig:usecase1}
\end{figure*}

\subsection{Qualitative Experiments} 
\label{sec:qual}

We also demonstrate the use and effectiveness of \drx in two use cases. The first one is through a step-by-step process, as presented in Figure~\ref{fig:usecase1}. We use the Diabetes dataset to solve a regression problem with PCA as the DR technique. We choose an interpretable DR technique, in order to be able to compare \drx to the ground truth weights of PCA. The second one concerns using DR techniques for visualization.

\subsubsection{Use Case: Regression}
\label{sec:ucr}
We divide the dataset into a train set (80\%) and a test set (20\%). After training PCA, we reduce the dimensions of both the train and test sets from 10 to 8. In this manner, we preserve 95\% of the original data. Then, we train a Ridge regression model, which serves as the decision-making mechanism. Trained and evaluated on the reduced datasets, the model achieves a MAE score of 42.08. 

We then choose a random reduced instance $x_i' = [-0.73, -0.86,$ $-0.07, -0.88, 0.17, -0.07, -0.1, -0.06]$ to explain the prediction of the model. The prediction of ridge for this instance is 140.03. Ridge regression, because it is intrinsically interpretable, can provide a set of coefficients. Because our reduced set has 8 dimensions, we also have 8 coefficients $coef = [8.73, -72.23, 28.48, 74.1, -27.07, -55.25,$ $-15.1, 25.95]$. $coef$ presents the global weights of ridge, while $coef_{x_i'} = coef\times x_i' = [-6.38, 62.08, -1.9, -65.04, -4.51, 3.85, 1.47, -1.51]$ showcases the local interpretation for Ridge's prediction regarding the examined instance $x_i'$.

\begin{align}
\label{pca:1}
    x_i' = (x_i-mean) \cdot weights^T\\
\label{pca:2}
    x_i = x_i' \cdot weights + mean
\end{align}

\begin{table}[ht]
\caption{Contribution of features for each one of the $8$ components from PCA}
\centering
\begin{tabular}{ccccccccc}
\hline
\textbf{}    & \textbf{$C_1$}     & \textbf{$C_2$}    & \textbf{$C_3$}    & \textbf{$C_4$}     & \textbf{$C_5$}     & \textbf{$C_6$}   & \textbf{$C_7$}     & \textbf{$C_8$}     \\ \hline
$F_1$        & 0.1 & -0.35 & \textbf{-0.82} & -0.18 & -0.37 & -0.08 & -0.09 & -0.03 \\
$F_2$        & \textbf{0.96} & 0.24 & -0.01 & -0.03 & 0.06 & 0.01 & 0.03 & 0.09 \\ 
$F_3$        & 0.05 & -0.3 & 0.18 & 0.29 & -0.13 & -0.2 & -0.49 & \textbf{0.7} \\
$F_4$        & 0.1 & -0.34 & -0.19 & \textbf{0.57} & \textbf{0.59} & -0.26 & -0.02 & -0.32  \\
$F_5$        & 0.04 & -0.36 & 0.1 & -0.53 & 0.39 & 0.02 & -0.01 & 0.1   \\
$F_6$        & 0.05 & -0.23 & 0.13 & -0.41 & 0.23 & 0.03 & -0.35 & -0.14 \\
$F_7$        & -0.11 & 0.11 & -0.29 & -0.15 & 0.47 & 0.16 & 0.23 & 0.49  \\
$F_8$        & 0.12 & -0.26 & 0.29 & -0.16 & -0.18 & -0.12 & -0.15 & -0.32 \\
$F_9$        & 0.09 & \textbf{-0.46} & 0.22 & -0.0 & -0.18 & -0.23 & \textbf{0.74} & 0.17  \\
$F_{10}$     & 0.1 & -0.37 & 0.06 & 0.25 & -0.06 &\textbf{ 0.89} & 0.02 & -0.02 \\ \hline
\end{tabular}
\label{tab:PCAFeatureContribution}
\end{table}

In PCA, an instance is reduced based on Eq.~\ref{pca:1}, while it is inverse transformed through Eq.~\ref{pca:2}. Influenced by this inverse transformation process of PCA, we will propagate this local interpretation $coef_{x_i'}$ to the original space to produce $coef_{x_i}$, through \(coef_{x_i} = coef_{x_i'} \cdot weights + mean\) (Eq.4). PCA's coefficients are presented in Table~\ref{tab:PCAFeatureContribution}.

\begin{figure}[ht]
\centerline{\includegraphics[width=0.45\textwidth]{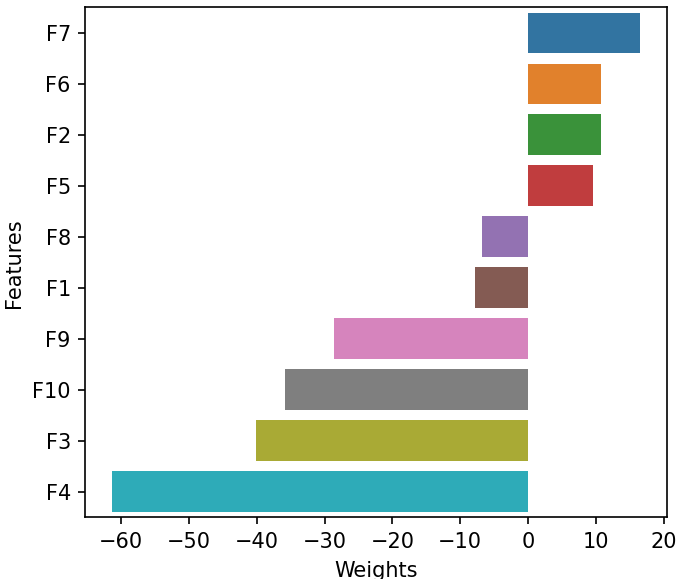}}
\caption{Local interpretation of $x_i$'s prediction in the original feature space}
\label{localInterpretation}
\end{figure}

By doing this multiplication, we have $coef_{x_i}$, which is the local interpretation of the Ridge model's prediction of the random instance in the original feature space. Therefore, $coef_{x_i} = [-6.38, 62.08, -1.9,$ $-65.04, -4.51, 3.85, 1.47, -1.51]$. We also present this information in Figure~\ref{localInterpretation}.

$F_4$ is the most important feature in the ridge's prediction. As a result, we perform the following test. We decrease the value of the 4$^{th}$ feature from $-0.121$ to $0.0021$ (the feature's mean value). Because the importance is negative, we anticipate that the final prediction will be increased. We transform the modified instance with PCA before querying Ridge for another prediction. Ridge predicts a score of $146.86$, which is higher than the initial prediction of $140.03$. Hence, our hypothesis is correct.

\begin{table}[ht]
\caption{Contribution of features for each one of the $8$ components from \drx}
\centering
\begin{tabular}{ccccccccc}
\hline
\textbf{}    & \textbf{$C_1$}     & \textbf{$C_2$}    & \textbf{$C_3$}    & \textbf{$C_4$}     & \textbf{$C_5$}     & \textbf{$C_6$}   & \textbf{$C_7$}     & \textbf{$C_8$}     \\ \hline
$F_1$   & 0.1   & -0.35 & \textbf{-0.82} & -0.18 & -0.37 & -0.08 & -0.09 & -0.03 \\ 
$F_2$   & \textbf{0.96}  & 0.24  & -0.01 & -0.03 & 0.06  & 0.01  & 0.03  & 0.09 \\ 
$F_3$   & 0.05  & -0.3  & 0.18  & 0.29  & -0.13 & -0.2  & -0.49 & \textbf{0.7} \\ 
$F_4$   & 0.1   & -0.34 & -0.19 & \textbf{0.57}  & \textbf{0.59}  & -0.26 & -0.02 & -0.32 \\ 
$F_5$   & 0.04  & -0.36 & 0.1   & -0.53 & 0.39  & 0.02  & -0.01 & 0.1 \\ 
$F_6$   & 0.05  & -0.23 & 0.13  & -0.41 & 0.23  & 0.03  & -0.35 & -0.14 \\ 
$F_7$   & -0.11 & 0.11  & -0.29 & -0.15 & 0.47  & 0.16  & 0.23  & 0.49 \\ 
$F_8$   & 0.12  & -0.26 & 0.29  & -0.16 & -0.18 & -0.12 & -0.15 & -0.32 \\ 
$F_9$   & 0.09  & \textbf{-0.46} & 0.22  & -0.0  & -0.18 & -0.23 & \textbf{0.74}  & 0.17    \\ 
$F_{10}$&  0.1  & -0.37 & 0.06  & 0.25  & -0.06 & \textbf{0.89}  & 0.02  & -0.02 \\ \hline
\end{tabular}%
\label{tab:FeatureContribution}
\end{table}

Nevertheless, a lot of DR techniques, mostly non-linear, cannot provide information about the $weights$ out of the box. Hence, a technique like \drx can provide this kind of interpretation. Consider the PCA as a black box, and therefore we do not have the weights of Table~\ref{tab:PCAFeatureContribution}. We are using \drx to provide a local explanation of how the reduced representation of the examined instance was influenced by the original features. The parameters we select in this example are: AA is True, NG is KNNs, K is 150. By initializing \drx, we request an interpretation. The output of \drx is visible in Table~\ref{tab:FeatureContribution}.

\begin{figure}[ht]
\centering
\begin{minipage}{.45\textwidth}
    \centerline{\includegraphics[width=\textwidth]{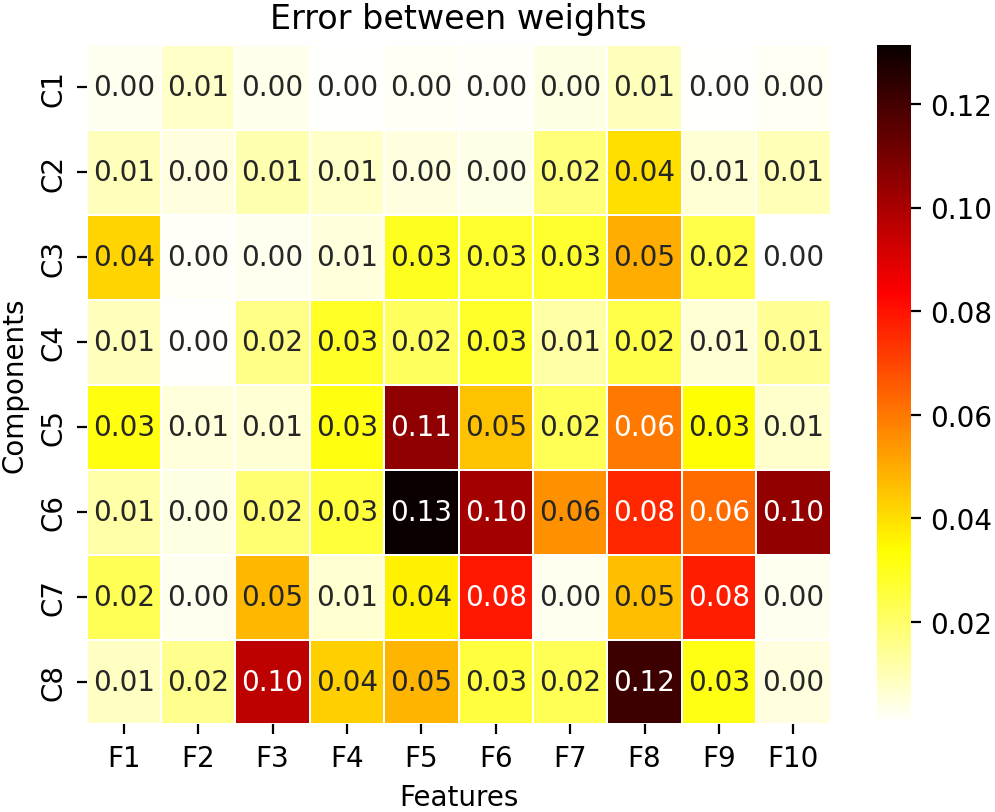}}
    \caption{Weights difference between PCA and \drx. For visualization purposes, all values have been multiplied by 1000}
    \label{fig:heatmap}
\end{minipage}%
\hfill
\begin{minipage}{.45\textwidth}
    \centerline{\includegraphics[width=\textwidth]{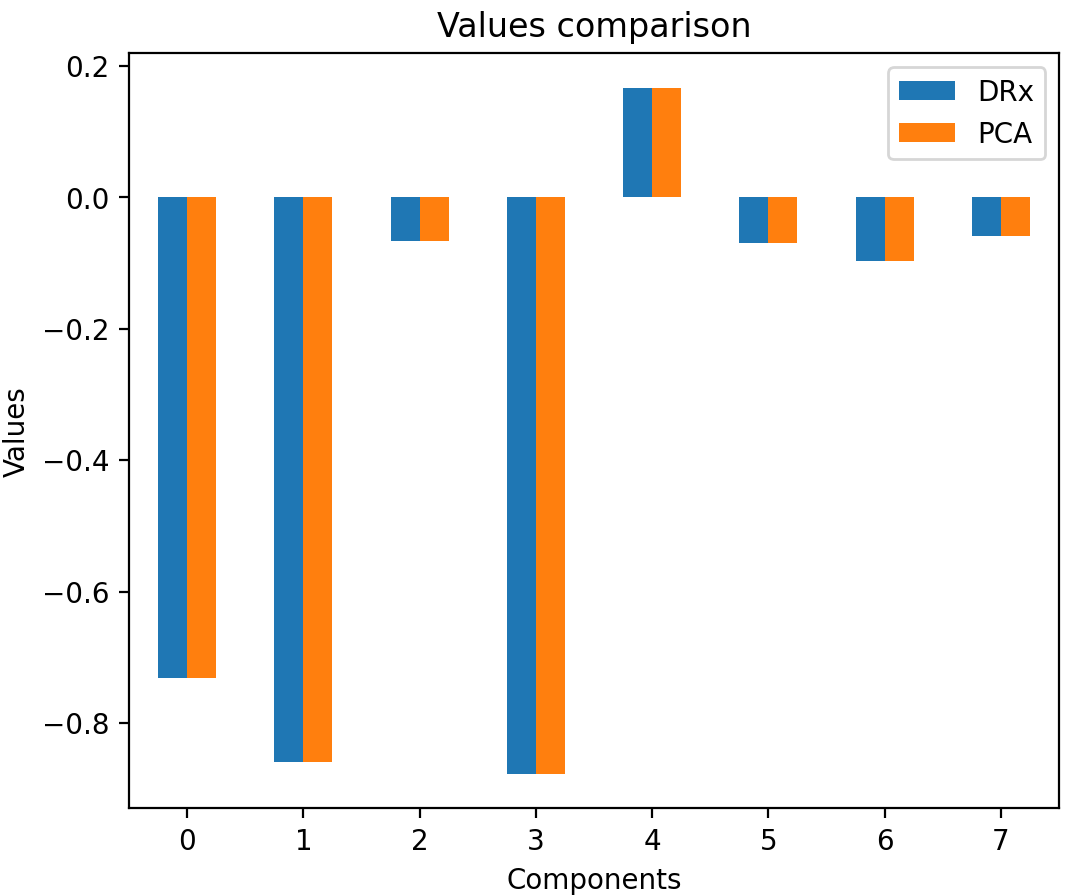}}
    \caption{Instance difference between PCA and \drx}
    \label{fig:15}
\end{minipage}
\end{figure}

Furthermore, we provide the \textit{weights difference} between PCA and \drx weights. The error in terms of ED is $0.0001$. The weight differences can also be seen in Figure~\ref{fig:heatmap}, where the differences are extremely small. It is clear that the weights are nearly identical. As a result, even if PCA could not provide this interpretation intrinsically, we could have computed such weights accurately with \drx. The last comparison is to reduce the original instance using the PCA and \drx weights. As seen in Figure~\ref{fig:15}, the reduced representations of the instance are almost identical. Indeed, the ED \textit{instance difference} is $5.54e-05$.

Researchers and end-users who want to evaluate and implement \drx in their solutions should use the provided metrics for parameter tuning. When the DR technique is uninterpretable, the parameters should be tuned and the \textit{instance difference} should be measured. On the other hand, if the DR is interpretable, both the \textit{instance} and the \textit{weights differences} should be used to tune the parameters. 

\subsubsection{Use Case: Visualization}

DR techniques are frequently utilized to facilitate data visualization by reducing data to two or three dimensions. In this use case, we are using the same dataset, but we reduce the dimensions to two to visualize them using RBF KPCA. 

\begin{figure}[ht]
\centering
\begin{minipage}{.45\textwidth}
    \centerline{\includegraphics[width=\textwidth]{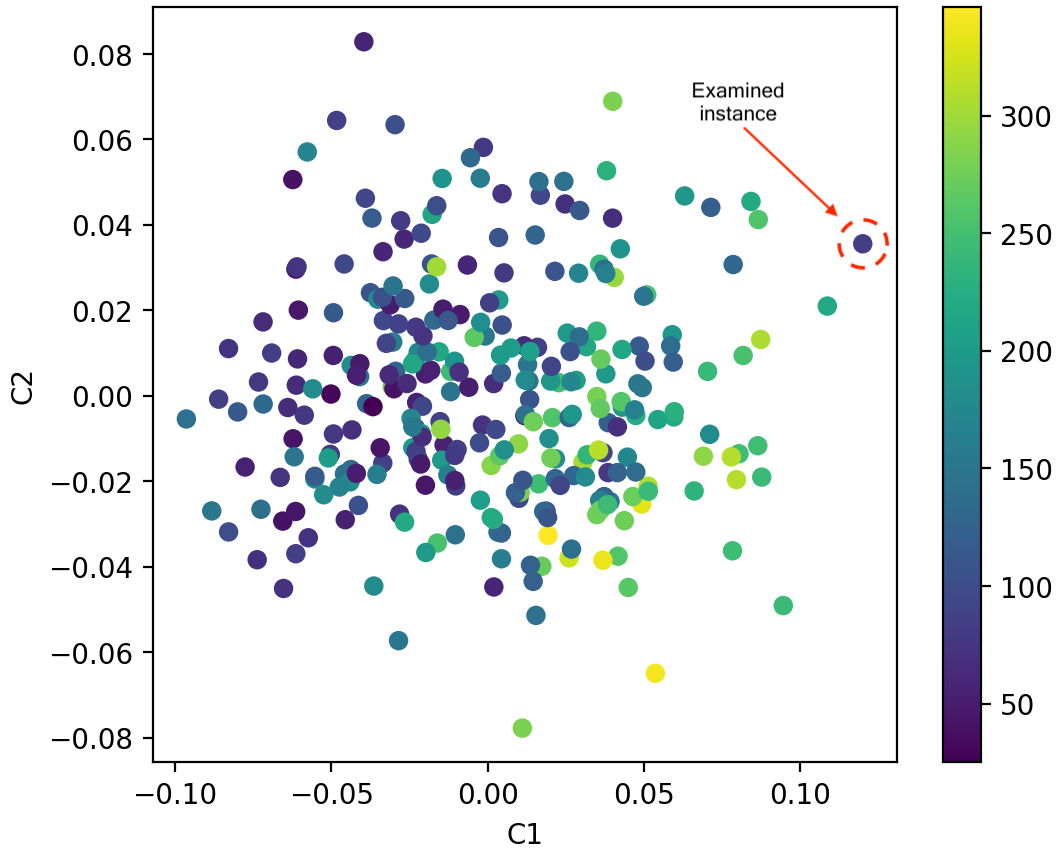}}
    \caption{Visualized data after KPCA reduction to 2 dimensions}
    \label{fig:outlier}
\end{minipage}%
\hfill
\begin{minipage}{.45\textwidth}
    \centerline{\includegraphics[width=\textwidth]{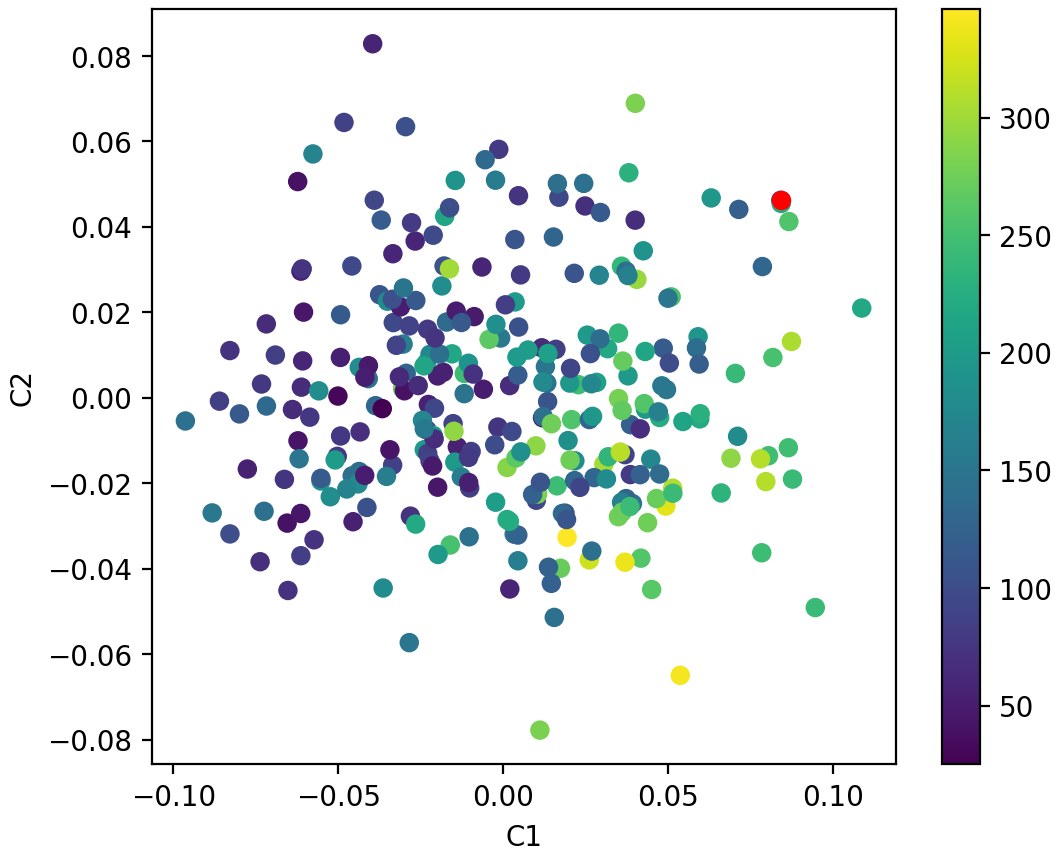}}
    \caption{Visualized data after KPCA reduction to 2 dimensions with tweaked $x_e$ instance}
    \label{fig:17}
\end{minipage}
\end{figure}

In Figure~\ref{fig:outlier}, we visualize the data. The dataset concerns a regression problem, and the color map represents the target value. It is visible that on the right part of the figure the target values tend to be higher. However, there is a particular point (circled in red) that seems to be an outlier; it has a high target value while lying far from samples with similar targets. Employing \drx we can investigate why this point has a high value on the first component ($C_1$).

This instance's original representation is $x_e = [0.01, 0.05, 0.03,$ $-0.0, 0.15, 0.2, -0.06, 0.19, 0.02, 0.07]$. The reduced representation through KPCA is $x_e'=[0.12, 0.04]$. Figure~\ref{fig:outlier} reveals that the abnormal value of $x_e$ concerns $C_1$'s $0.12$ value. Using \drx we acquire the weights representing the influence of the input features to each component. The weights regarding the component we examine, $C_1$, are $[0.08, 0.08, 0.14, 0.11, 0.16, 0.14, -0.11, 0.21, 0.16, 0.14]$. The 8$^{th}$ feature appears to have the highest impact on the value of $C_1$. We repeat the experiment as earlier. We change the value of $F_8$ from $0.19$ to $0$ (the feature's mean value), and then we reduce the instance again through KPCA.

The modified instance's representation is $x_{e\_tweaked}'=[0.08,$ $0.05]$. In Figure~\ref{fig:17}, we notice how the instance got closer to the center, and it is no longer considered an outlier. If we wanted to use this data for training, we should either exclude this point from the training set or investigate whether its target value is incorrect.

\section{Conclusions}
\label{sec:con}
We introduced \drx in this paper, a technique for providing local interpretations for DR techniques. We were able to investigate and determine which original features contributed the most to a specific reduced component by combining neighboring data with the predictions of a linear model. Experiments were conducted to support the effectiveness of \drx. In three different datasets, we compared the results of \drx to those of its global variant, GXDR, using two different metrics. \drx outperforms GXDR in every experiment. Furthermore, we investigated how datasets with larger dimension spaces affect \drx performance. 

We also performed a scalability analysis to assess \drx's response time. While for smaller datasets applying \drx is fast, in larger datasets the response time exceeds the one second. Therefore, in order to utilize \drx in an online, production application further optimization of the process can be applied. For example, parallelization of the training process of the different linear models, or precomputed neighborhoods.

Regarding the qualitative experiments, we presented two use cases. The first use case concerns the usefulness of \drx in interpreting the results of a regression model trained on data reduced by a DR technique. The second involves inspecting individual points in a plot that appear to be anomalous after being projected by a DR technique.

In the future, we can investigate new parameters for \drx to improve performance. We intend to adapt \drx to be applicable to DR techniques that cannot reduce new instances, such as t-SNE. The application of \drx to image and textual datasets should be investigated as well. Ways to improve the evaluation of \drx performance, inspired by the evaluation metrics used in the IML research on supervised tasks, will also be examined. Finally, a user study can be conducted to determine how effective a technique like \drx may be for (non)expert users.

\section*{Acknowledgments}
The research work was supported by the Hellenic Foundation for Research and Innovation (H.F.R.I.) under the ``First Call for H.F.R.I. Research Projects to support Faculty members and Researchers and the procurement of high-cost research equipment grant'' (Project Number: 514)

\bibliographystyle{unsrt}  

\end{document}